\documentclass[10pt,twocolumn,letterpaper]{article}

\usepackage[pagenumbers]{cvpr} 

\pdfoutput=1

\usepackage{graphicx}
\usepackage{tabularx,booktabs}
\usepackage{adjustbox}
\usepackage{amsmath}
\usepackage{amssymb}
\usepackage{booktabs}

\usepackage[capitalize]{cleveref}
\crefname{section}{Sec.}{Secs.}
\Crefname{section}{Section}{Sections}
\Crefname{table}{Table}{Tables}
\crefname{table}{Tab.}{Tabs.}

\def\cvprPaperID{65} 
\def\confName{CVPR}
\def\confYear{2022}

\begin{document}

\title{Comparison of CoModGANs, LaMa and GLIDE for Art Inpainting\\
Completing M.C Escher's Print Gallery}

\author{Lucia Cipolina-Kun\\
ML Collective and \\ 
University of Bristol. UK \\
{\tt\small lucia.kun@bristol.ac.uk}
\and
Simone Caenazzo\\
Riskcare Ltd., London\\
\and Gaston Mazzei\\
University Paris-Saclay. CNRS\\
LISN, VENISE team,Orsay,France.
}

\maketitle

\begin{abstract}
   Digital art restoration has benefited from inpainting models to correct the degradation or missing sections of a painting. This work compares three current state-of-the art models for inpainting of large missing regions. We provide qualitative and quantitative comparison of the performance by CoModGANs, LaMa and GLIDE in inpainting of blurry and missing sections of images. We use Escher's incomplete painting \textit{Print Gallery} as our test study since it presents several of the challenges commonly present in restorative inpainting.
\end{abstract}



\section{Introduction}
\label{sec:intro}


Artworks and images are part of our cultural heritage, but have a tendency to deteriorate over time. Inpainting is a restoration technique that has been applied traditionally to restore or complete the missing or damaged sections in a way that the restorative work passes unnoticed. In cases where the missing region is of considerable size, this task becomes delicate as the aim is to fill-in the area with content that ensembles well with the painting, whilst also fitting the painter's style and historical period. 

\smallskip
With the recent development of Machine Learning techniques, new inpainting models are available to the Cultural Heritage restorers. However, at present only few models are developed specifically with artwork restoration in mind. The training of these models requires dataset of images in the counts of thousands, a laborious and resource-intensive task \emph{per-se}. The traditional solution is to fine-tune these models and re-train them with images similar to the restored piece; this is typically also a challenge, as it can be difficult to provide large sets of examples of relevant artwork. 

\smallskip
Some examples of inpainting models specifically developed for art reconstruction include the works of Guptal \etal \cite{Guptal} and Amiri and Messinger \cite{Amiri}, which both propose models derived from computer vision inpainting and extended to the art domain. Note that in both works domain experts were used to evaluate model performance, as an acknowledgment of the specific difficulty of evaluating inpainting in the specific context of art.

Our work aims to extend the current literature and provide an evaluation on how current state-of-the-art inpainting models can be used in an art restoration context. We offer a qualitative and quantitative comparison of three models developed for the inpainting of large missing sections, namely CoModGANs \cite{comodgan}, LaMa \cite{LaMa} and GLIDE \cite{GLIDE2021}. It is worth noting how none of these models was developed specifically for art reconstruction; however, given their versatility and simplicity, our aim is to show the context on which each of them can be successfully used as a restorative tool. 

\smallskip
In order to stress-test the models in a challenging territory, we selected M.C Escher's lithography, \emph{Print Gallery}, as a test case. This work contains an entire missing region at the center where different semantic contents blend, thus being an excellent test case for inpainting models. Additionally, we compare the performance of each model in other well-known artworks, like the Ecce Homo by Elias Garcia Martinez and Escher's "Bird-Fish" used to highlight the weak and strengths of each model under different settings.


\section{Inpainting Methods for Large Regions}

The focus of our comparison is Computer Vision models developed specifically for the inpainting of large missing regions. In such contexts, the unmasked regions typically provide little information to guide the model towards the right choice of content, thus presenting additional challenges. Additionally, the larger the output required from the model, the more evident effects like pixelation and content mismatches can be. The three models selected are currently the state-of-the-art models for large-mask inpainting tasks. 

\textbf{CoModGAN}. The model \textit{Large Scale Image Completion via Co-Modulated Generative Adversarial Networks (CoModGANs)}, implements a modulation of the unconditional image vectors into the traditional Generative Adversarial Networks to generate content consistent with the image's semantics. It is based on the \textit{StyleGAN} family of models \cite{StyleGan}, which allows to control salient features or styles of an image. To enhance performance on large-mask inpainting tasks, the model was trained using randomized large masks over the training datasets. A limitation of the current model distribution is the relatively low resolution required for input images: the model was trained on images of 512x512 size, requiring any other input image to match such size (thus potentially lowering the resolution of the overall output) when using the model. The model was trained on \textit{Places2} \cite{Zhou2018-places},  CelebA-HQ \cite{liu2015-celeb} and COCO-Stuff \cite{COCO} datasets, making it versatile for a wide type of objects.

\textbf{LaMa}. The model \textit{Resolution-robust Large Mask Inpainting with Fourier Convolutions} was designed with large regions in mind as well. It is a simple deterministic Pix2Pix-like model \cite{Pix2Pix} with segmentation-based perceptual loss and a ResNet-like architecture with fast Fourier convolutions instead of the  \textit{StyleGan} logic. The strength of the model is to target regular patterns in an image to repeat them across the masked region. The results of the model largely depend on the presence of regularities on the area surrounding the masked region.  For example, in images with tiles, bricks and windows \textit{surrounding} the mask. As an advantage over the others, this model is able to work with a higher resolution of 2048x2048. The model was trained only on two datasets, \textit{Places2} and \textit{CelebA-HQ}. 

\textbf{GLIDE}. The model \textit{Guided Language-to-Image Diffusion for Generation and Editing} is a multimodal diffusion model with text guidance. Diffusion models work similarly to upsampling models: the generator net is trained by progressively adding noise to an image and the learning objective is to revert the noise process, generating a de-noised image back. An additional component is the \textit{text-guided} module, which allows the user to guide the image generation process by inserting a text prompt that acts like an additional constraint to the model. This prompt allows for virtually infinite possibilities in the number of outputs generated, while also avoiding the inconvenience of fine-tuning large models, as is the case of CoModGAN and LaMa. Additional model parameters such as the \textit{guidance scale} and \textit{temperature} allow the user to control the mix of conditional and unconditional outputs. An ablation study of GLIDE's parameters is presented on the supplemental material. Resolution-wise, the released version of GLIDE accepts image inputs as large as 6Kx6K pixels; however, it then down-samples inputs to 64x64 for memory optimization and on the last stage it up-samples them back to 256x256, which is its final output resolution. The upsampling process, together with its training are the key to producing its claimed photorealistic quality. The model version released by OpenAI was trained on a fil-
tered dataset excluding human figures from the MS-COCO \cite{MSCOCO}  dataset for images the and CLIP's dataset for text \cite{CLIP}.
%
\begin{table}[!htbp]
  \raggedright
  \small
  \begin{tabular}{lc@{} lc@{} lc@{} lc@{}}
    \toprule
    Model    & Type && Input size && Output Size  \\
    \midrule
    CoModGANs & StyleGan && 512x512 && 512x512 \\
    LaMa      & Fourier Conv && 2048x2048  && 2048x2048\\
    GLIDE     & Text guided diff && 6000x6000  && 256x256\\ 
    \bottomrule
  \end{tabular}
  \caption{Comparison of model type, input and output sizes across models.}
  \label{tab:comparison}
\end{table}



\section{M.C. Escher's Print Gallery}

\begin{figure}[ht!]
  \centering
   \includegraphics[width=\linewidth]{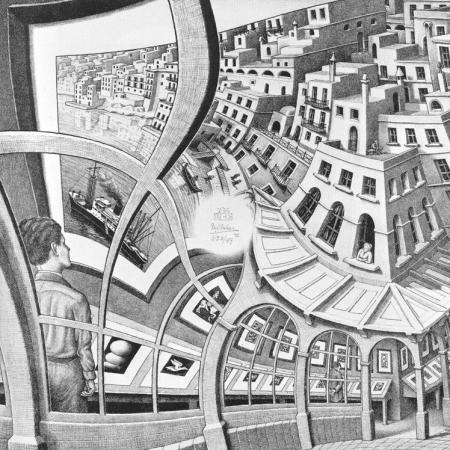}

   \caption{M.C. Escher’s lithography \emph{Print Gallery} (\emph{Prentententoonstelling}), 1956. Image from Wikipedia Commons.}
   \label{fig:PrintGallery}
\end{figure}

The artwork chosen for the present model testing exercise is \textit{Print Gallery} (original title: \emph{Prentententoonstelling}), made in 1956 by the Dutch artist M. C. Escher. Figure \ref{fig:PrintGallery} presents the original lithography, portraying a man that observes a painting in a gallery; the painting, in turn, portrays a gallery in the waterfront of the Grand Harbour of Valletta in Malta. 

\textit{Print Gallery} is a peculiar work for several reasons:
\begin{itemize}
    \item It features a so-called \emph{Droste effect} (i.e. a roto-homothecy): the man stands in a gallery which is eventually portrayed again in the painting he is observing, creating a theoretical infinite loop;
    \item The painting is embedded in a spiral-like structure, clearly evident in the twisting of buildings, columns and other elements;
    \item Lastly, Escher did not complete the center of the painting, instead only adding his signature in the resulting blank - to this day, there is no definitive explanation for his choice.
\end{itemize}

The seemingly incomplete nature of the painting is arguably the main reason for the notoriety of \emph{Print Gallery} among the artistic and mathematical community. Due to its challenging nature, several mathematicians and artist have attempted to complete it  \cite{Hofstadter1979, Lenstra}. In \cite{Lenstra}, Lenstra and de Smit present a class of exponential (conformal) complex maps \cite{HandbookNumAnalysis} that share a similar shape with the spiral-like structure in the original painting. Such maps provide a bridge between a normal, undistorted space and the twisted space of the painting. Their work provides the mathematical foundation that we leverage upon in this present paper, together with Machine Learning techniques, to complete the center of the \emph{original Print Gallery}. In particular, we show how the conformal map formulation can be used to pave the way for Computer Vision techniques - the performance of which we aim to compare as the main objective of our work. 

\subsection{Unrolling From Warped to Straight}\label{sec:canonical_form}

It is worth noting that any attempt to apply Computer Vision/inpainting techniques \emph{directly} on the blank of \emph{Print Gallery} is faced with two main complications:

\begin{enumerate}
    \item The painting, as described in the previous section, features a significant amount of twisting and rotation - Machine Learning models are, in general, not equipped to deal with extreme transformations in the sample image, since they are not equivariant to rotations, scaling and generally warping of images  \cite{Goodfellow2016};
    \item The size of objects to be completed in the center is very small in relation to the rest of the painting, once again creating challenges for any model trying to understand the sample image context.
\end{enumerate}

The exponential maps described in \cite{Lenstra} provides a solution, in that the twisted space in Print Gallery can be deconstructed into eight \emph{straightened pictures in the Euclidean space} - each of which features an incomplete area in the shape of a spiral. This set of eight pictures are individually inpainted to complete the center. 

\smallskip
The two equations below provide the mappings to first translate Pirnt Gallery's warped space into the Euclidean space (obtaining the eight straight images) and then back from Euclidean to warped. Let $z = (x,y)$ be the coordinates of RGB pixels in the complex plane, with $(x,y)$ being standard Cartesian coordinates, and $T(z):\mathbb{C} \rightarrow \mathbb{C}$ be the following complex exponential map:

\begin{equation}
\label{eq:to-warped}
    T(z) = exp^{\alpha Ln(z)} 
\end{equation}

In the specific case of \emph{Print Gallery}, a suitable value of the constant is: $\alpha = \frac{2 \Pi i + Ln(256)}{2 \Pi i }$ \cite{Lenstra}. Note that Equation \ref{eq:to-warped} maps the straight Euclidean space into an \textit{approximation} for the twisted space featured in the original \emph{Print Gallery}. In order to map the twisted space into the straight one, we define the inverse map $T^{-1}$ as below:

\begin{equation}
\label{eq:to-straight}
    T^{-1}(z) = exp^{\frac{1}{\alpha} Ln(z)} 
\end{equation}

\begin{figure}[h]
\centering
\includegraphics[width=.5\textwidth]{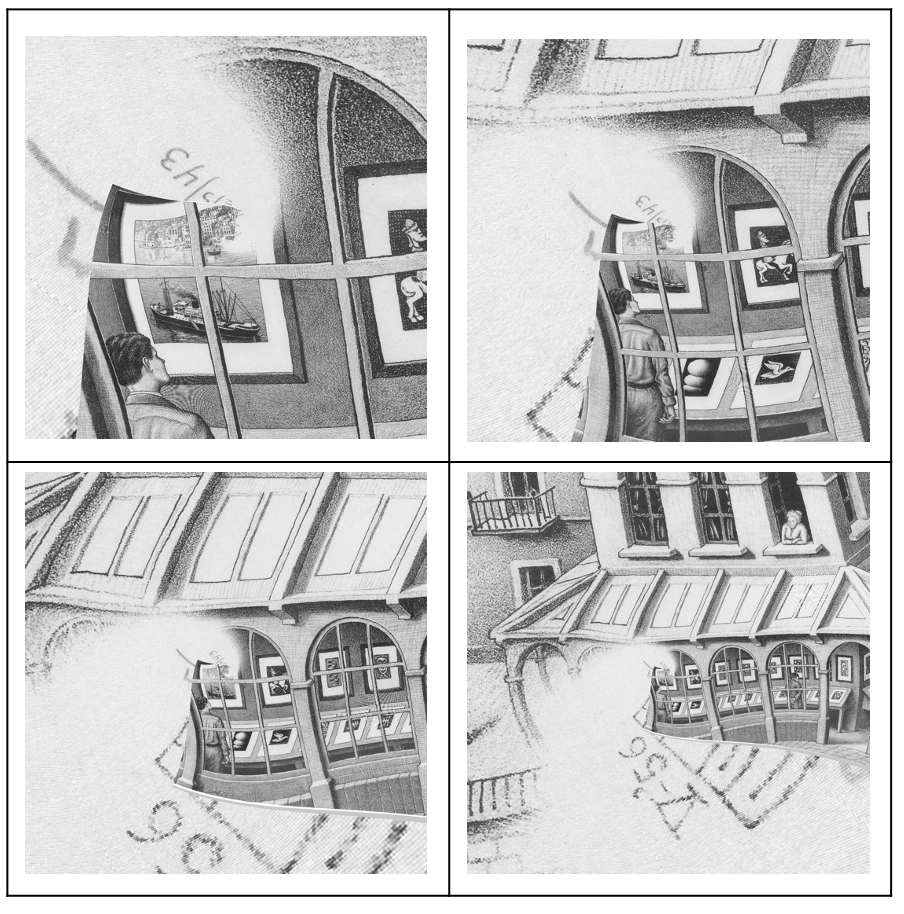}
\includegraphics[width=.5\textwidth]{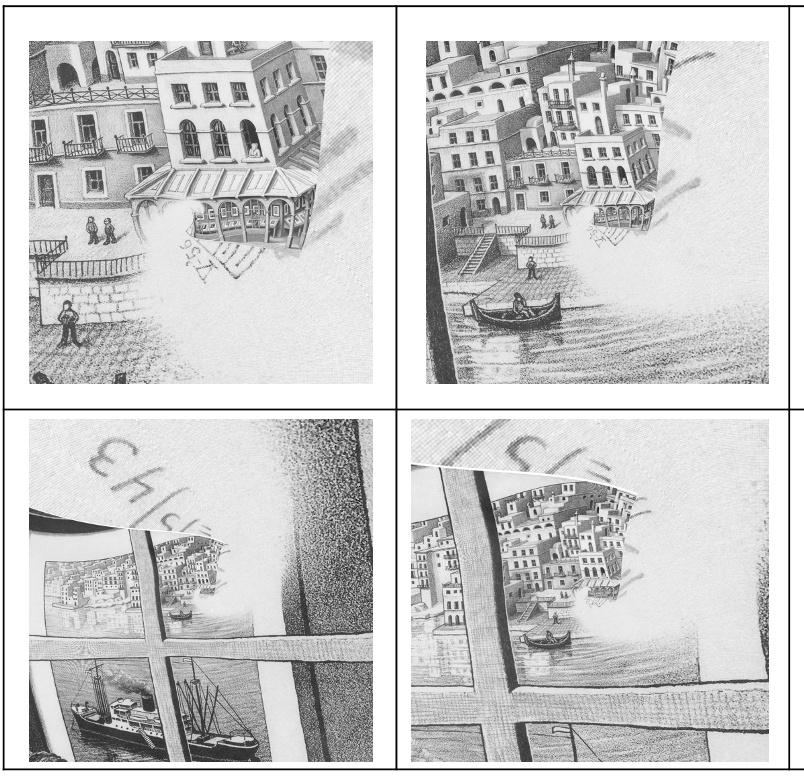}
\caption{M.C. Escher’s lithography converted into eight straight images. The blank center appears as a white spiral on each image.}\label{fig:4straight-b}
\end{figure}

Note that Equation \ref{eq:to-straight} describes a one-to-many mapping, as it is in fact periodic, with period $4^4 =256$. As a result of the periodicity of the map, we obtain a set of eight straight images from Print Gallery, (which are shown in Figure \ref{fig:4straight-b}), each of which is in relation to the next one via a zoom factor of 4.

\smallskip
The eight straight images obtained are used in the model comparison exercise as follows. First apply Equation \ref{eq:to-straight} to \emph{Print Gallery}, obtaining 8 straight sample images, second apply the tested models on the straight sample images, aiming to complete the spiral-shaped blank region, and lastly evaluate the performance of the models on each of the eight straight images. Summary metrics of our testings together with qualitative examples are presented in the following sections.  
  

\section{Model Comparison}

Assessing the quality of an image depends very much on its context and usage. In the case of digital art, while the technical correctness of a restoration is important, there is an increased importance on \emph{subjective} qualities of the restoration. We evaluated the three models using a group of subjective criteria such as: artistic consonance with the rest of the lithography;adherence to the painter's style and adherence of any new content to the historical period depicted in the artwork. Additionally, we compared model outputs using objective metrics traditionally used for \textit{no-reference} image quality assessment. 

\subsection{Qualitative Analysis}

As mentioned in Section \ref{sec:canonical_form}, we tested the three models on the inpainting of the straight images in \cref{fig:4straight-b}. For CoModGAN we used the demo provided\footnote{https://github.com/zsyzzsoft/co-mod-gan} with the \textit{Places 2} dataset. For LaMa, we used the demo provided with the \textit{high-quality} setting \footnote{https://cleanup.pictures/}. For GLIDE we used the Colab demo with a \textit{Guidance Scale} of 4 and the text prompt \textit{"Print Gallery"} \footnote{https://github.com/openai/glide-text2im/blob/main/notebooks/inpaint.ipynb}.

\begin{figure}[ht]
\includegraphics[width=0.5\textwidth]{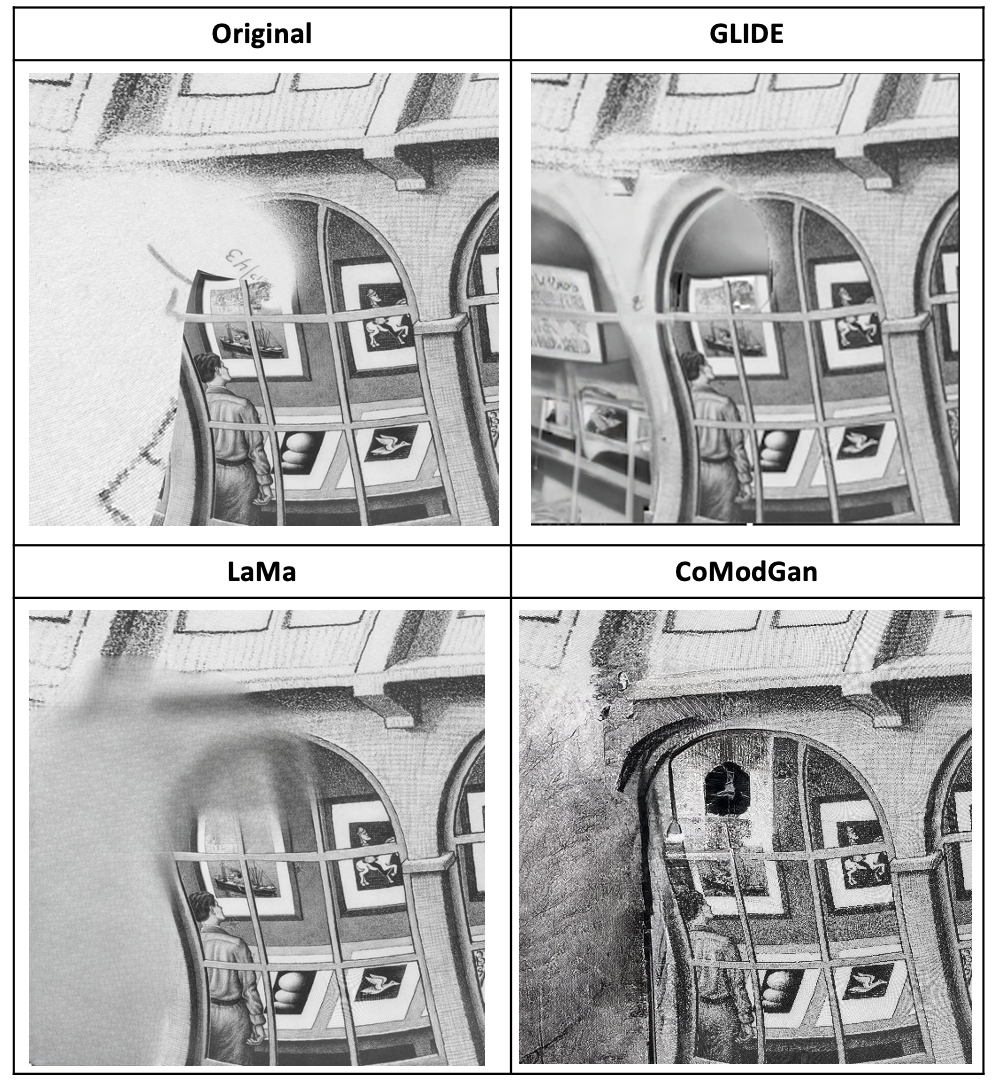}
  \centering
\caption{Example of model output for the same mask. The white area on the upper left image is the masked region.}
\label{fig:straight-completed} 
\end{figure}

\cref{fig:straight-completed} shows an example output for each of the three models. The top-left image shows the target masked image. Note that the mask is placed in the left-most border of the image, requiring the model to do inpainting as well as outpainting. This is an important observation, as CoModGAN and LaMa are models not natively suited for outpainting.

\smallskip
We now outline findings of the qualitative analysis in the form of conclusions.

\medskip
\textit{Conclusion 1. The model output is significantly determined by the placement of the mask.}

\medskip
The three models evaluated are heavily dependant on the pixels surrounding the masked region. GLIDE and CoModGANs have a higher context awareness than LaMa. Besides the context, GLIDE is highly influenced by the prompt and other tunable parameters. An ablation study of GLIDE's parameters is presented on the Appendix and the supplementary material.

\medskip
\textit{Conclusion 2. GLIDE's output is determined by the prompt, the seed and the guidance scale parameter, which determines the degree at which the prompt affects the output. For LaMa and CoModGANs, the only way to improve the output image is by performing costly fine-tuning.}
\medskip

Due to its multimodality, GLIDE can produce, in theory, an infinite number of outputs for the same mask, solely by changing the seed and the text prompt. This allows the user to rank the outputs or handpick the best inpainting solution for the context. The other models give a single output option per masked region, and thus, are more sensitive to the mask definition.

\medskip
\textit{Conclusion 3. GLIDE is superior in outpainting (extrapolation) tasks when compared to LaMa and CoModGANs.}
\medskip

LaMa and CoModGANs are models developed for inpainting, this is, their output is primarily based on the information content read from the surrounding pixels of the mask. However, in outpainting, the mask extends beyond the borders of the image which leads the model with no surrounding information to work with. 
The images on the bottom show that LaMa and CoModGANs under-perform on outpainting tasks. This is in line with expectations, since none of them were developed specifically for outpainting. 

\medskip
\textit{Conclusion 4. GLIDE has a higher output variance, often producing uncanny objects.}

\medskip
Different from GAN models, GLIDE was not trained using a discriminator net, which is used to avoid the production of unrealistic artifacts. GLIDE on the other hand, is mostly text-guided, and as result, it produces a wide variance of surrealistic objects. In the artistic arena this diversity can be beneficial depending on the use case. The diversity of GLIDE's output will be further analyzed on \cref{subsec:GLIDE_variance}.

\subsection{Detailed analysis of each model}

This section analyses the results from each individual model in more detail. We present cases of both good performance and failures of each, with the aim of showing  the aim is to show that each of these models specializes on different domains. To summarise, LaMa performs exceptionally well on image colors with well-defined patterns, CoModGANS is best suited for human faces and landscapes. As for GLIDE, while seems to be an all-terrain model, even capable of performing outpainting, its public release was filtered to not produce human figures. Additional comparison is presented in the supplementary material, where figures comparing the same failure cases across models are presented.

\cref{fig:comodgan} below shows two examples of CoModGANs runs with the masked region boxed in red. The left image shows the limitations of the model on a simple outpainting task, where the natural expectation would have been for it to follow the color pattern. The image on the right shows instead a setting where the model performs very well as the model correctly learned to reproduce the buildings surrounding the mask. While the content generated is correct from a visual point of view, it is not  in line with the painter's style or the historical period of the painting, as the CoModGANs model has been trained on the modern (Places2) dataset. The way to shift the generation into a more suitable content is to fine-tune the net, which requires building a dataset of related artworks in the count of thousands, which is usually not available. 

\begin{figure}[ht]
\includegraphics[width=0.48\textwidth]{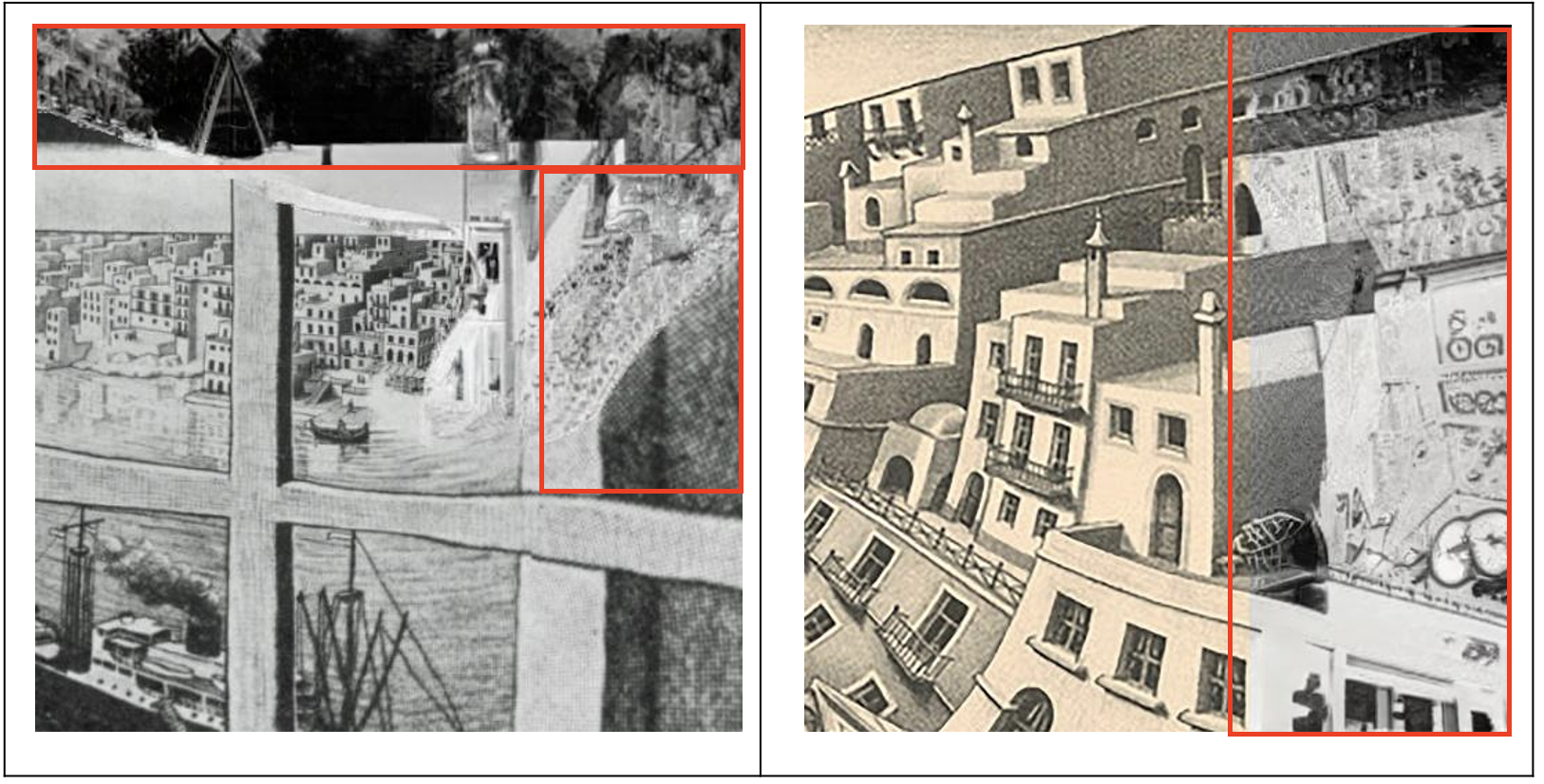}
  \centering
\caption{Example outputs of CoModGANs. The masked region is boxed in red. Note the graffiti painting produced by CoModGANs on the right image.}
\label{fig:comodgan} 
\end{figure}

\smallskip
\cref{fig:lama} below shows two examples of LaMa runs, again masked regions are shown boxed in red. The image on the left shows how the model fails on outpainting of images; in this particular case, the sample image presents a high degree of pixelation, making the recognition task harder. The output on the right shows a correct output, where the model correctly identifies and mimics the pattern present in the surroundings of the mask. While the produced content is correct, the output still shows a certain degree of blurriness and pixelation.

\begin{figure}[ht]
\includegraphics[width=0.48\textwidth]{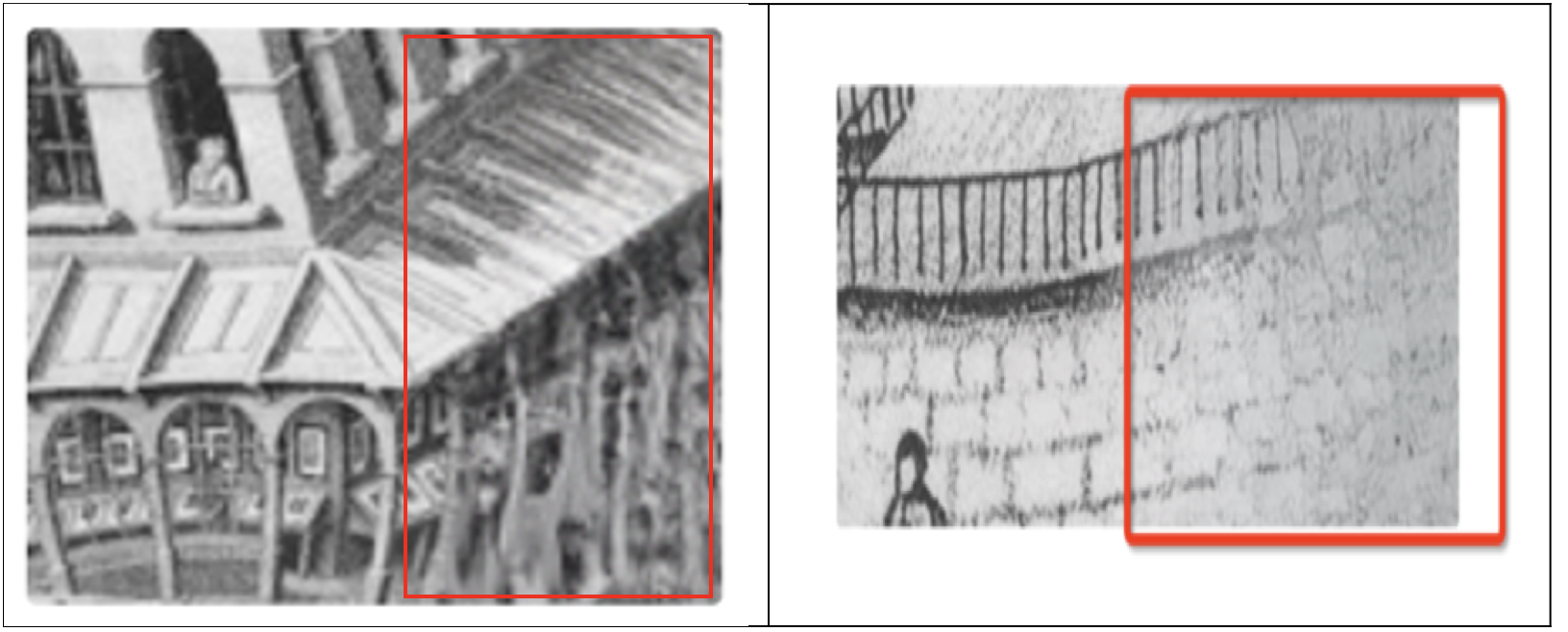}
  \centering
\caption{Example outputs from LaMa. Masked region boxed in red.}
\label{fig:lama} 
\end{figure}

\cref{fig:glide} below shows two failure cases of GLIDE. On the left image, the model tries to mimic the human figure and fails, producing additional inconsistent details. This is likely due to the fact that GLIDE's training dataset does not contain humans, as a design choice. The image on the right shows a failure as a consequence of the model's output variance which is further analysed on \cref{subsec:GLIDE_variance}. We can see how the model produces unrealistic objects, which have no resemblance with a particular object on its training set. This could be explained by the fact that the model does not contain a discriminator network, as the output is only guided by the cosine similarity with the text prompt.

\begin{figure}[ht]
\includegraphics[width=0.48 \textwidth]{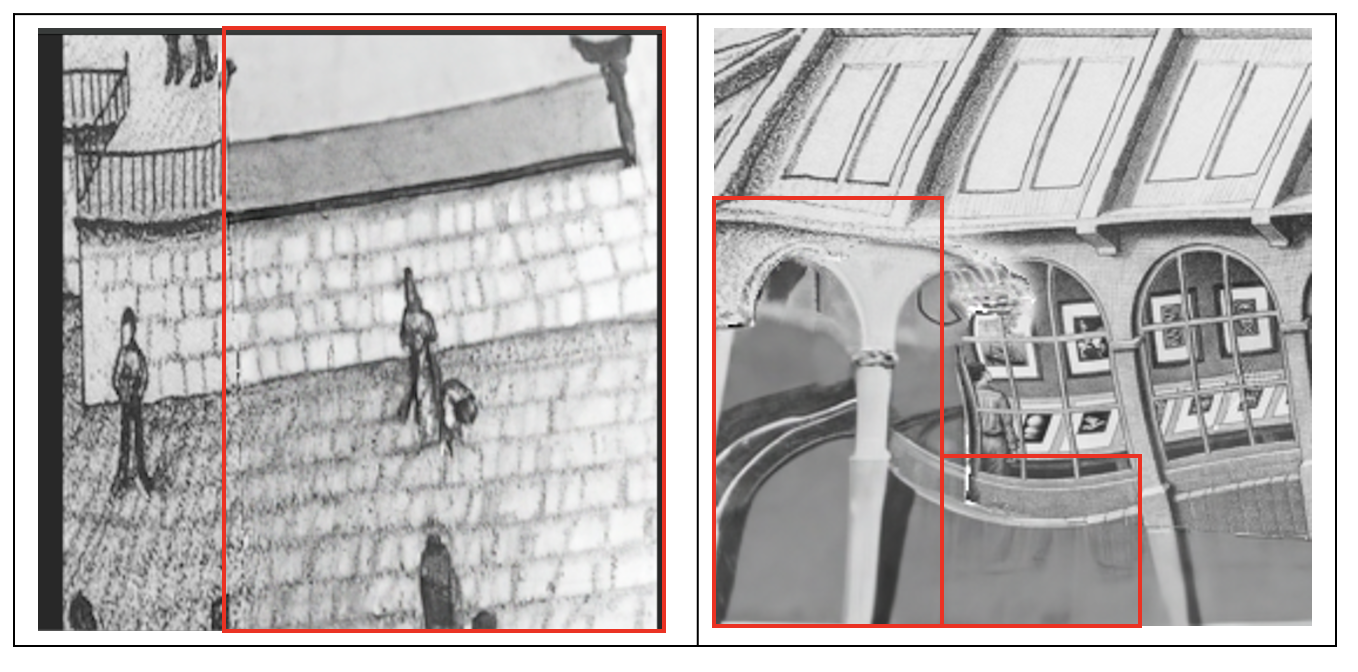}
  \centering
\caption{Example outputs from GLIDE. Masked sections boxed in red. Guiding Scale of 5}
\label{fig:glide} 
\end{figure}

\subsection{Analysis of GLIDE's Output Diversity}
\label{subsec:GLIDE_variance}
As explained before, GLIDE's distinctive feature is its multimodality, it takes as input a masked image with a text prompt and produces a (theoretically) infinite supply of inpainting options. This creates the problem of image selection; it is not clear a priory, how many batches of images are needed to find the best inpainting option and additionally, there is no selection metric provided with the model. 

\smallskip
An example of the diversity of GLIDE's output is shown below. We generated samples for the same mask, prompt and seed. We can see that the output is very dissimilar among the images selected and in a sense uncanny with the expectations for an Escher painting. Note that here we analyze dissimilarity over the content created, and not on image quality. 

\begin{figure}[ht]
\includegraphics[width=\columnwidth]{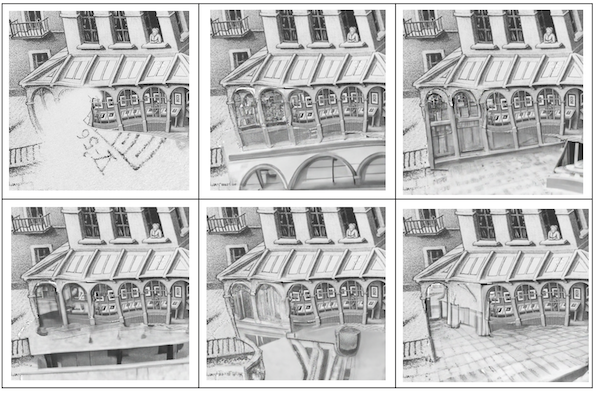}
\caption{Examples of inpainted images generated by GLIDE for the caption \emph{"a gallery with arches wooden windows and arcades and floors with tiles"} and a Guiding Scale of 5. The white area in the top-left image is the masked region.}
\label{fig:GLIDE_outputs} 
\end{figure}

To measure the diversity of the inpainted content created in an objective way, we calculated the CLIP score over 250 random samples of the top-left image in \ref{fig:GLIDE_outputs}, using the same mask and prompt \footnote{The prompt used is "A man looks at a painting of Malta behind the windows of a gallery"}. The CLIP score measures the cosine similarity between the text prompt and the output image  \cite{CLIP}, a higher text prompt means the content created resembles better the passed prompt. While the Coefficient of Variation of the CLIP score is only $3.62 \%$, in visual terms, this variation translates into very significantly distinctive content. Additional outputs are shown on the supplementary material. 


\subsection{Analysis on Different Paintings}
This section shows the performance of the models under alternative settings other than the eight straight images obtained from Print Gallery \footnote{Additional examples and a longer analysis is presented on the Supplementary Material section.}.  The main conclusion is that each of the analyzed models has been developed and trained for a specific use-case and there is no model that outperforms the others across the board, when it comes to qualitative assessment. 

\begin{figure}[ht]
\includegraphics[width=\columnwidth]{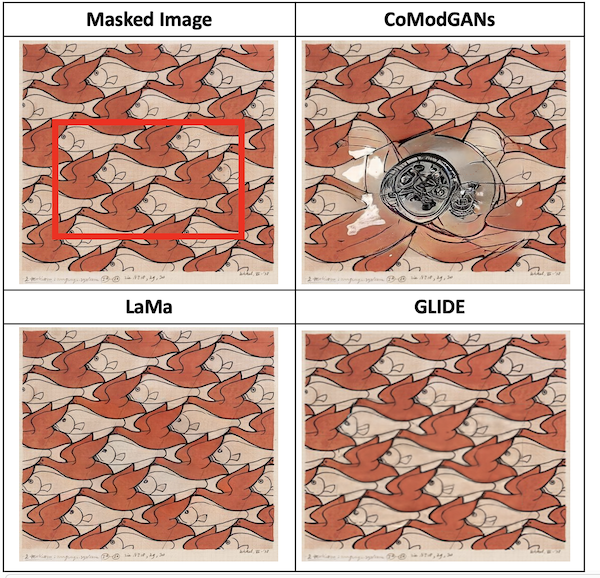}
  \centering
\caption{M.C. Escher's Bird-Fish painting. 1938. Comparison of performance of the three models over a regular painting. The masked region is the entire square area delimited in red. Image reproduced under WikiMedia Commons.}
\label{fig:fish_birds} 
\end{figure}

\smallskip
The Figure \ref{fig:fish_birds} shows a painting with clear color patterns where LaMa's performance is the strongest as expected for a Fourier-based model. In fact, the only difference with the original image is the detail of the reconstruction of the eyes.  GLIDE shows good results however, LaMa's output is at 2048x2048 while GLIDE is only able to provide a quality of 256x256 \footnote{GLIDE's prompt used is simply "pattern" and the Guidance scale is five. A low guidance scale helps the model to favor the image's semantics over the text prompt}.

\begin{figure}[ht]
\includegraphics[width=0.48\textwidth]{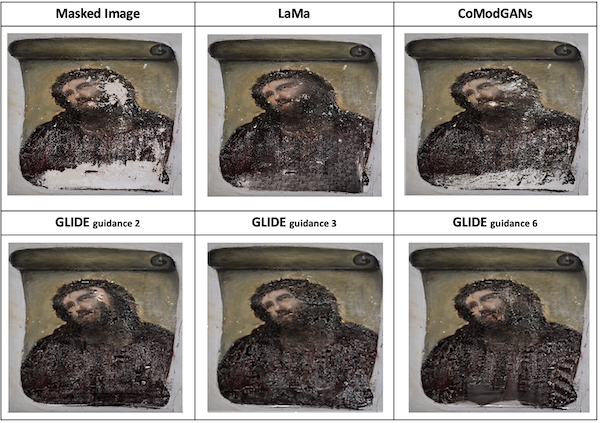}
  \centering
\caption{Sample outputs from different models on the Ecce Homo fresco by Elias Garcia Martinez. GLIDE's outputs are presented for Guidance Scales of two, three and six. Image reproduced under WikiMedia Commons.}
\label{fig:ecce_homo} 
\end{figure}

The image on Figure \ref{fig:ecce_homo} shows the limitations of GANS-based models on digital restoration. In particular CoModGANs is trying to blend the masked region with the neighboring colors, missing the context, as is a feature of the localized convolution of GANs.  While LaMa succeeds on the face part, it fails on the lower part of the image. GLIDE's outputs varies with the Guidance Scale parameter, however, none of the outputs is able to recognize the feature of a human face as its distributed version has been restricted to not produce humans \footnote{GLIDE prompt used is "a man staring like Jesus with shirt red and black stripes".}. 

\subsection{Quantitative Metrics}

We used three different metrics to provide a quantitative comparison of the models' outputs as shown in \cref{tab:QuantResults}. The selected metrics are commonly used in the field of \textit{no-reference} image quality assessment, where the quality of an image is determined without using any target image for comparison. In our case, each image was evaluated as a stand-alone output. The model \textit{Koniq} produces a score by comparing the input image against the largest dataset of image quality up to date \cite{koniq}.  The model \textit{BRISQUE} reports a score using a Support Vector Regression trained on an annotated image dataset with known distortions \cite{Mittal2012}; such dataset is, however, biased towards landscape pictures. Lastly, we used the \textit{DOM} \cite{DOM} model which gives a score based on the sharpness of gray images. 

\smallskip
To obtain a diverse sample of images, we tested the models across the eight straight images in Figure \ref{fig:4straight-b}, which contain large regions of inpainting and outpainting challenges. We created 50 different random masks on each model and used the same mask across models. The use of 50 masks is justified by an ANOVA test presented on the Appendix in \cref{Appendix1}.

Analysing Table \ref{tab:QuantResults} we can see that in all cases GLIDE shows a superior performance, except for the DOM score, which shows GLIDE almost matching with CoModGANs on sharpness \footnote{GLIDE was run with a Guidance Score of 5 and Upsample Temperature of 0.997. The Supplemental material shows further analysis of GLIDE on the relationship between its parameters and the DOM score}. The good performance of GLIDE on the Koniq and BRISQUE scores are in line with the recent literature showing that, in general, diffusion models beat GANs on image synthesis \cite{DiffusionBeatsGANs}. This result can be explained by several factors. First the upsampling module present on GLIDE's acts similarly to a denoising feature creating a uniform density of pixels across an image. 

\smallskip
\textit{Conclusion 5. GLIDE presents superior performance on blurriness and deformation while not on image sharpness. However its performance is dependent upon the parameter tuning. }

\begin{table}[!htbp]
  \centering
  \begin{tabular}{lc@{} lc@{} lc@{} lc@{}}
    \toprule
    Method    & Koniq $\uparrow$ && Brisque $\downarrow$ && Dom $\uparrow$ \\
    \midrule
    CoModGANs & 36.12 && 43.37  && 1.05 \\
    LaMa      & 38.76 && 42.38  && 1.10\\
    GLIDE     & 41.61 && 7.94   && 1.04 \\
    \bottomrule
  \end{tabular}
  \caption{Average values for each metric. A higher Koniq score is better, a lower Brisque score is better and a higher DOM (edge sharpness) score is better.}
  \label{tab:QuantResults}
\end{table}

\section{Print Gallery Inpainting Result}
Figure \ref{fig:result} below displays the result of Print Gallery completed by performing three steps. First we applied Eq. \ref{eq:to-straight} obtaining the eight straight images in Figure \ref{fig:4straight-b}, second we completed the missing region of each using GLIDE \footnote{The parameters used and additional details of the completion process can be found on the supplementary material}, and lastly we combined the eight straight images as in \cref{eq:to-warped} to obtain back Print Gallery. 

\begin{figure}[h!]
\centering
\includegraphics[width=\columnwidth]{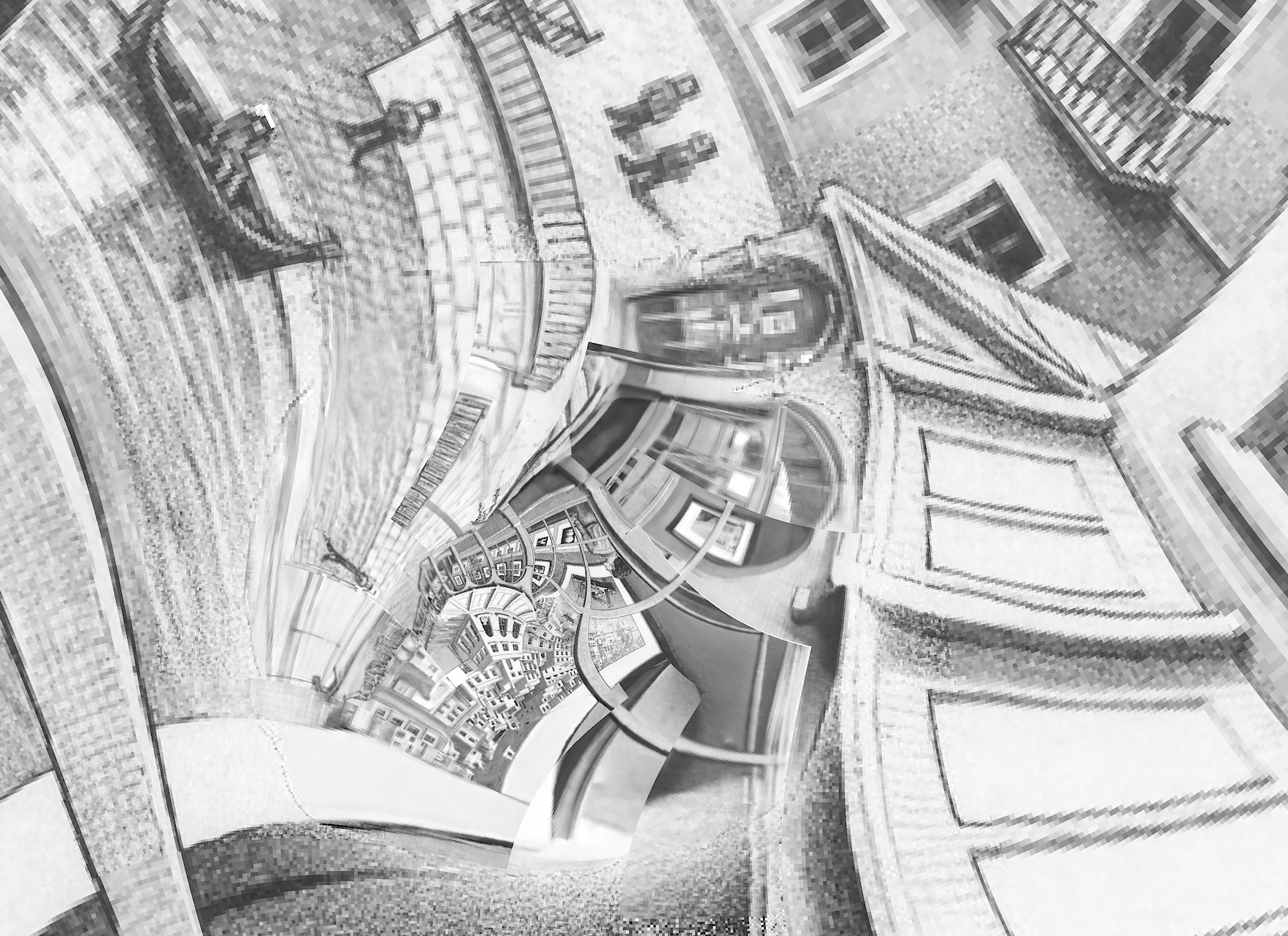}
\caption{Detail of the completed center using GLIDE. In the center-bottom it shows the repetition of the original rotated by 157 degrees.}
\label{fig:center}
\end{figure}

\smallskip
Figure \ref{fig:center} displays a zoom-in of the completed center. It is noticeable some mismatch between the boundaries of the warped straight images, this is due to the difference in Escher's original lithography and the parametrized mappings applied in \cref{eq:to-straight} and \cref{eq:to-warped}. To correct for this, a future direction is presented on Section \ref{sec:future_work}.
Note how the inpainted region is very small and rotated for any inpainting model to be used out of the box (i.e. without any fine tuning or passing to the Euclidean plane)\footnote{The supplementary material shows an alternative completions made by hand by professional artists }. Additionally, as a consequence of the one-to-many mapping in \cref{eq:to-straight} the center presents an homothecy of Print Gallery itself, rotated by 157 degrees.

\begin{figure}[h!]
\includegraphics[width=\columnwidth]{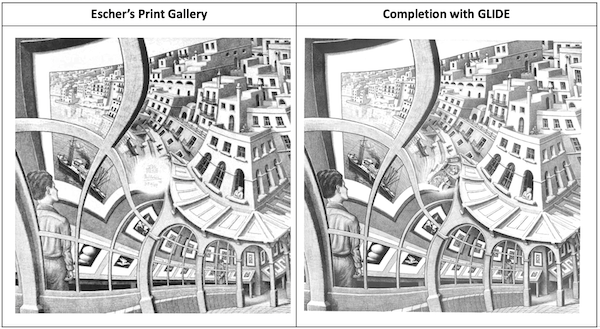}
\caption{Comparison between original Print Gallery and our completion using GLIDE}
\label{fig:result}
\end{figure}

\section{Conclusions}

We have provided a quantitative and qualitative analysis of three of the current state-of-the-art models for inpainting on large masks. By using a particularly challenging setting, comprised of a mixture of inpainting and outpainting modalities over different images, we have obtained test-case results for each model's strengths and weaknesses. GLIDE appears to be superior to LaMa and CoModGANs on outpainting tasks and it is benefited from an upsampling module obtaining photorealistic quality. Additionally, GLIDE provides the user with alternative completions for a given mask and prompt, which can be beneficial on artistic settings and allows one to calibrate the output result without the costly fine-tuning required by the other two methods. However, GLIDE's output diversity can also lead to unrealistic outputs and thus, requires human discretion to select the best fit. We have shown how . According to expectations, LaMa was shown to be superior in pattern-replication tasks, and it has the best resolution output across all models. As for CoModGANs, similar to the family of StyleGANs models, it shows best performance on big masks over human faces and landscapes, since it was specifically trained on them, while GLIDE's dataset filtered out human images.

\section{Future Work}
\label{sec:future_work}

As mentioned in Section \ref{sec:canonical_form}, the formulas in \cref{eq:to-straight} and \cref{eq:to-warped} have been used to translate the original Print Gallery lithography into eight straight images. This is, however, an imperfect process due to the natural differences between a hand-made process and any attempt to parametrize it with closed-form formulas. To address this difference, we propose to project Escher's Print Gallery onto the conformal map space, for example using Thin Plate Splines (TPS) \cite{duchon1977}.  

\section{Appendix}
\label{Appendix1}

To test for the statistical significance of the 50 means on table \ref{tab:QuantResults}, we performed a one-way ANOVA test summarized below on Table \ref{tab:ANOVA}. We can conclude that the average values presented on table  are statistically different across all metrics, notwithstanding DOM which presents similar results for CoModGANs and GLIDE.

\begin{table}[!htbp]
  \centering
  \begin{tabular}{lc@{} lc@{} lc@{} lc@{}}
    \toprule
    Method    & Fvalue && Fcrit && RH0 \\
    \midrule
    Koniq     & 9.23  && 3.05  && yes \\
    Brisque   & 255.6 && 3.05 && yes\\
    DOM       & 74.17 && 3.05   && yes \\
    \bottomrule
  \end{tabular}
  \caption{Results of the ANOVA test performed over the mean results of the image quality metrics}
  \label{tab:ANOVA}
\end{table}

\section{Acknowledgements}
We thank the anonymous reviewers for their incisive comments that were most useful in revising this paper. Additionally, the authors would like to thank Rosanne Liu, Cris Luengo, Roman Suvorov, Vaisakh M, Andrea Panizza, Alejandro Cabrera, Jim Schmitz, Pablo Samuel Castro, Meire Fortunato, Pablo Sprechmann, Rick Anderson, Niranjan Krishna and Vahid Yazdanpanah.

\section{Supplementary Material}

\subsection{Further Qualitative Results}

\begin{figure}[!htbp]
\centering
\includegraphics[width=\columnwidth]{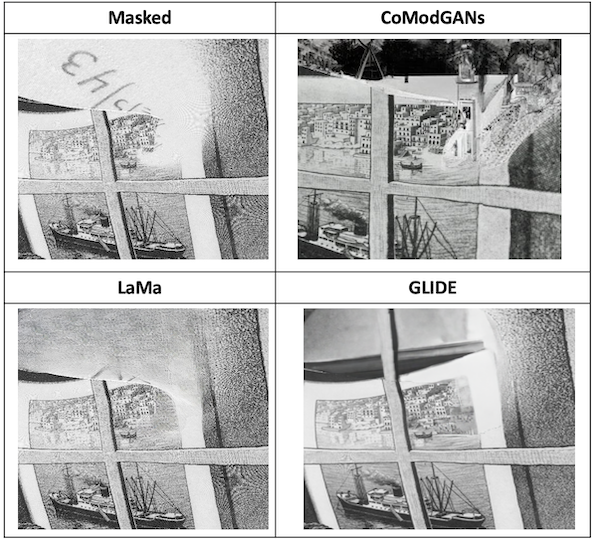}
\caption{Comparison of outpainting image across different models.GLIDE caption i s"window" and Guidance Scale of five.}\label{fig:comparison-window}
\bigbreak
\includegraphics[width=\columnwidth]{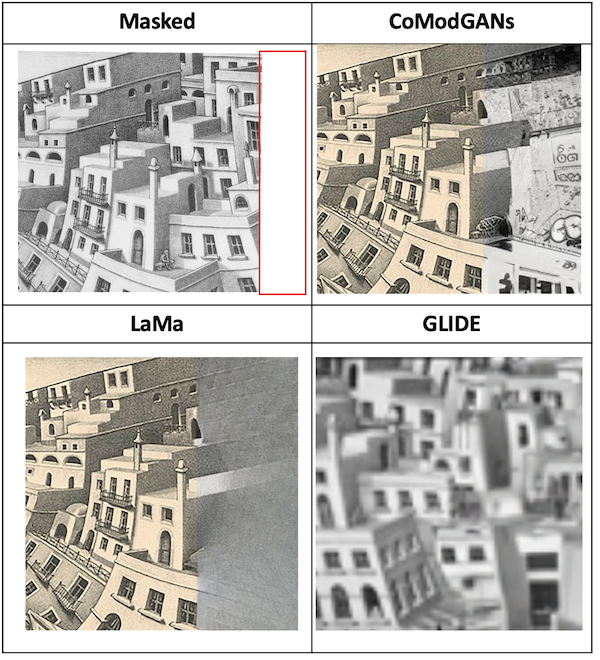}
\caption{Comparison of outpainting image across different models. GLIDE caption is "buildings" and Guidance Scale of five.}\label{fig:comparison-buildings}

\end{figure}

\begin{figure}[!htbp]
\includegraphics[width=\columnwidth]{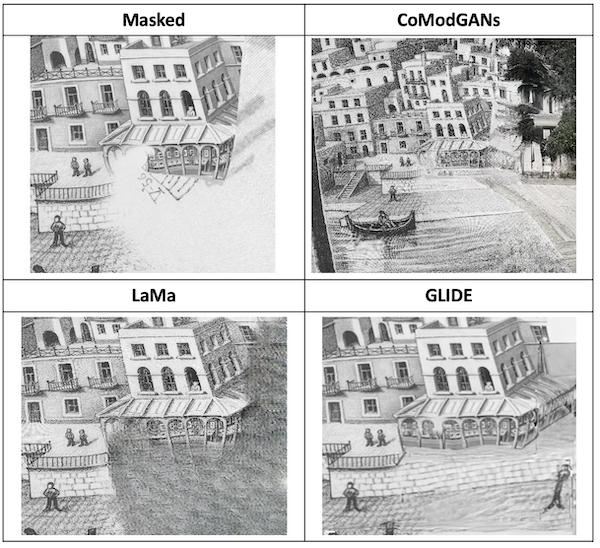}
\caption{Comparison of outpainting/inpainting image across different models. GLIDE caption is "buildings" and Guidance Scale of five.}\label{fig:comparison-7}
\end{figure}

The panels \ref{fig:comparison-window},\ref{fig:comparison-buildings},\ref{fig:comparison-7} present additional comparison of the 'failure' cases across different models. It is clear that LaMa and CoModGans are not suited for outpainting tasks. GLIDE on the other hand, performs well across inpainting and outpainting demands but suffers from having the worst resolution output at only 256x256.

\subsection{Additional Studies on GLIDE's Parameters}

\begin{figure}[!htbp]
\centering
\includegraphics[width=\columnwidth]{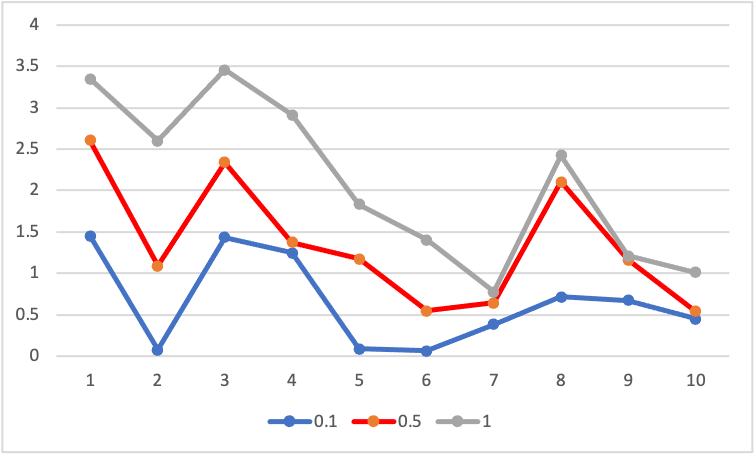}
\caption{Plot of BRISQUE values. The three categories represent the Upsampling Temperature and the horizontal axis represent the Guidance Scale }\label{fig:chart-BRISQUE}
\bigbreak
\includegraphics[width=\columnwidth]{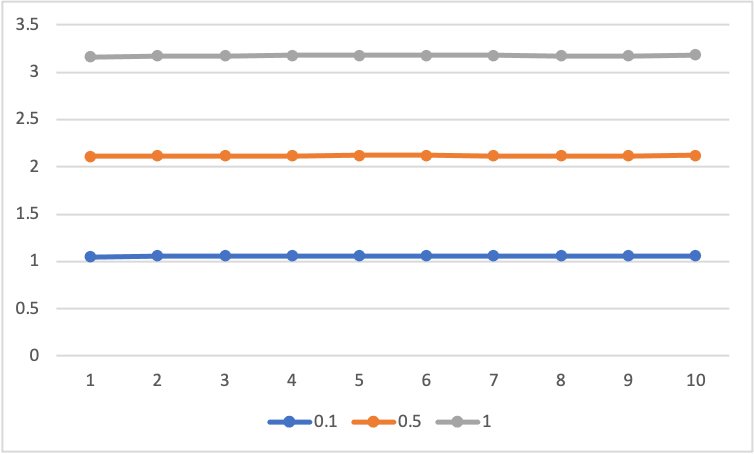}
\caption{Plot of DOM values. The three categories represent the Upsampling Temperature and the horizontal axis represent the Guidance Scale }\label{fig:chart-DOM}
\bigbreak
\includegraphics[width=\columnwidth]{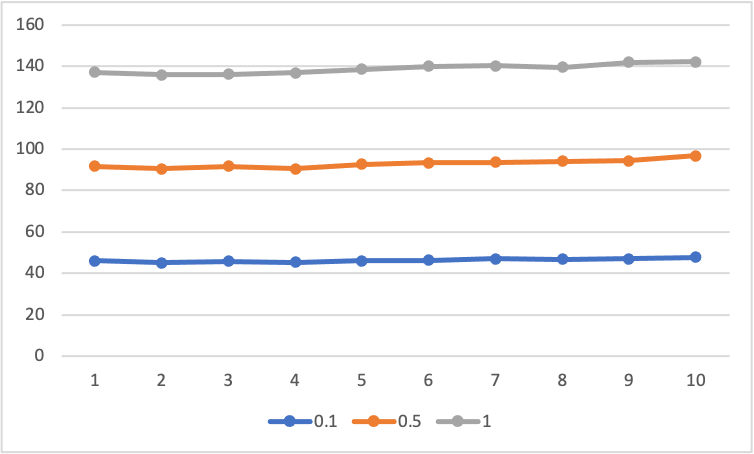}
\caption{Plot of KONIQ values. The three categories represent the Upsampling Temperature and the horizontal axis represent the Guidance Scale }\label{fig:chart-KONIQ}
\end{figure}

The panel in \ref{fig:chart-BRISQUE},\ref{fig:chart-DOM},\ref{fig:chart-KONIQ} shows the summary results of ablation studies performed on GLIDE. We tested the effect of changing the Guidance Scale (which controls the relationship between the prompt and the generated image) and the Upsampling Temperature (which controls the degree of upsampling) for the same image and same mask. The test was performed over 50 samples for each value \footnote{prompt: "a gallery with arches wooden windows and arcades floors with tiles"}. 

The results show a significant sensitivity of the model outputs to the parameters. As expected, a higher degree of upsampling improves results across all metrics. In fact, the recommended Upsampling Temperature is 0.997. The Guidance Scale controls the content, and thus, does not affect the sharpness or blurriness. In the case of BRISQUE, the sensitivity to the Guidance Scale might be explained by the degree of black color on the image, since this metric is sensitive to large black regions.

\begin{figure}[ht!]
\centering
\includegraphics[width=.5\textwidth]{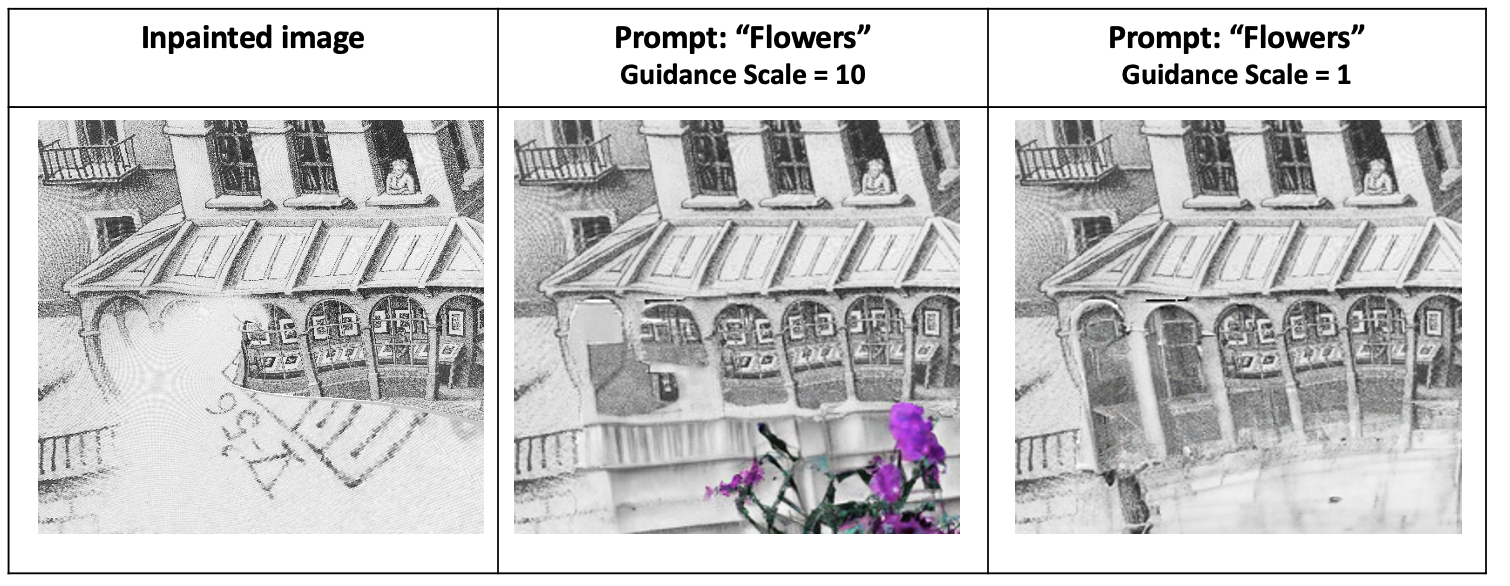}
\caption{GLIDE's outputs for different Guidance Scales and same prompt.}
\label{fig:GLIDE-flowers-comparison}
\end{figure}

Next we analyse the effect of the prompt on the inpainted content. As mentioned, GLIDE is a text-guided model where the incidence of the text is controlled by the parameter "Guidance Scale". Under a low Guidance Scale, GLIDE produces content in consonance with the surrounding objects, which is ideal for art inpatinting. On the contrary, a higher Guidance Scale gives more weight to the text prompt on the image generation. The effect of the Guidance Scale over the text guidance is shown on Figure \ref{fig:GLIDE-flowers-comparison} using a fixed seed and an Upsampling Temperature of 0.997.

\subsection{Detailed Analysis of Inpainted Print Gallery}

\begin{figure}[ht!]
\centering
\includegraphics[width=.5\textwidth]{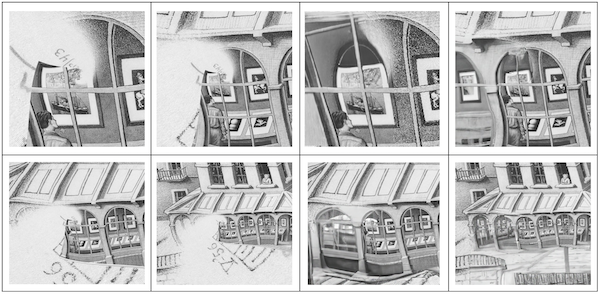}
\caption{Inpainting of the first four images composing the blank center. GLIDE with Guidance Scale of five and prompt "Escher".}\label{fig:8completed-a}
\bigbreak
\includegraphics[width=.5\textwidth]{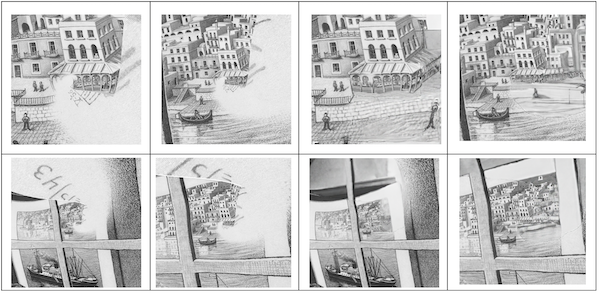}
\caption{Inpainting of the second four images composing the blank center. GLIDE with Guidance Scale of five and prompt "Escher".}\label{fig:8completed-b}
\end{figure}

The panel \ref{fig:8completed-a} shows a side-by-side comparison of the eight images composing the straight version of Print Gallery with their inpainted counterparts. On each panel, the four images on the right are the inpainted results of the the left images.

\begin{figure}[h]
  \includegraphics[width=.5\columnwidth]
    {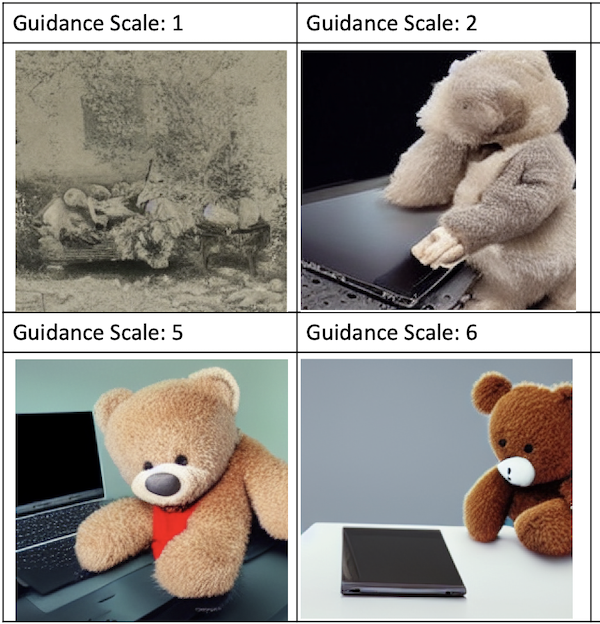}\hfill
  \includegraphics[width=.5\columnwidth]
    {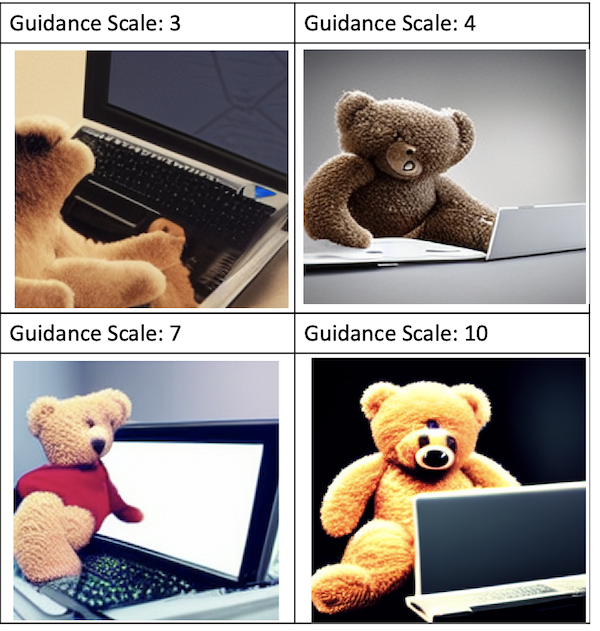}
  \caption{Different guidance scale settings over the same prompt.\textit{"a photograph of a teddy bear using a laptop 1080p 4k".}}
  \label{fig:bear2}
\end{figure}

On panels \ref{fig:bear2} we show a comparative analysis of GLIDE's output under different Guidance Scales, everything else equal. Results indicate that a Guidance Scale value of five results in the best object creation \footnote{Images by artist {@litevex} reproduced with permission}. The prompt used is: \textit{"a photograph of a teddy bear using a laptop 1080p 4k."}

\subsection{Analysis on Additional Paintings}

\begin{figure}[ht!]
  \includegraphics[width=1\columnwidth]
    {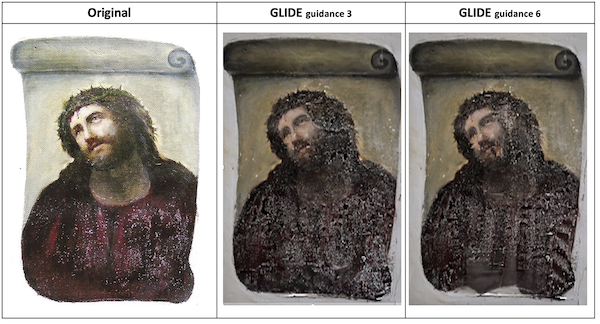}\hfill
  \includegraphics[width=1\columnwidth]
    {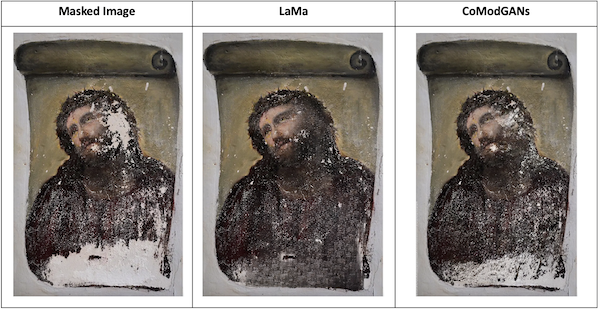}
  \caption{Different outputs for the Ecce Homo by Elias Garcia Martinez. Image from Wiki Commons. Glide was used with a Guidance Scale of three and six, where indicated.}
  \label{fig:Ecce-comparison1}
\end{figure}

This section provides further comparison of the model performance under different scenarios. First in \cref{fig:Ecce-comparison1} we provide a further analysis of the Ecce Homo by Elias Garcia Martinez from the main text. This fresco is a challenge since white patches from the degradation are visible all over the surface. In the case of CoModGANS, the model considers as a valid pattern the white areas appearing on the image's surface and tries to replicate them, resulting in a poor inpainting performance. LaMa repeats this behaviour but in a lesser degree, as noted, this model provides the best output resolution among all. Additionally, it performs comparatively well on the face since it has been trained on the faces database Celeb-H. GLIDE, as expected, it is not able to recognize a human face by design. However, the upsampling module produces a nitid result compared to others.

\subsubsection{Automatic Generation of Prompts}
\begin{figure}[ht!]
  \includegraphics[width=1\columnwidth]
    {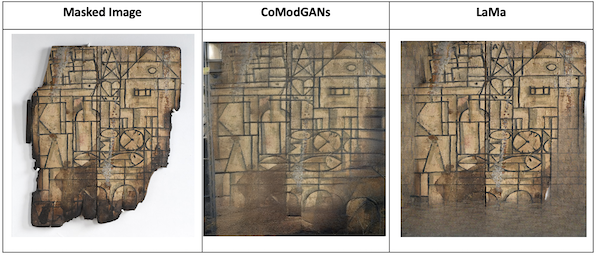}\hfill
  \includegraphics[width=1\columnwidth]
    {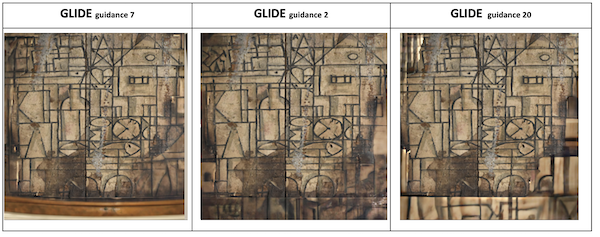}
  \caption{Inpainting outputs for the work of Torres-Garcia. Composicion constructiva (1932). GLIDE with different Guidance Scales and prompts generated by image-to-text bots. Image from Wiki Commons.}
  \label{fig:TG-comparison2}
\end{figure}

The \cref{fig:TG-comparison2} is a work by Torres-Garcia  "Composicion constructiva" (1932) \cite{TorresGarcia}. The piece was burnt at the Brazilian Museum of Modern art in 1981 and presents the traces of fire on the wood. This coloration creates a challenge for inpainting models as they deem the burnt area as a valid pattern to reproduce. We see that GLIDE is able to move away from the burnt colorization, either by ramping the Guidance Scale or by trimming the area with a clever prompt. For the left-bottom image, the prompt was generated using a image-to-text bot by the developers EleutherAI \footnote{www.eleuther.ai} , the generated prompt is "Paul Klee's rectangular piece of wood is a rare example of early Christian art.". The images with Guidance Scales two and 20 have used the simple prompt "patterns".

\subsubsection{Crowd-sourcing of Prompts By Experts}
\begin{figure}[h]
  \includegraphics[width=1\columnwidth]
    {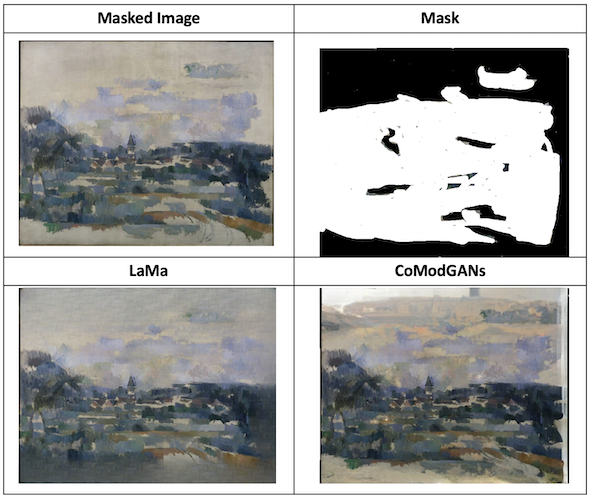}\hfill
  \includegraphics[width=1\columnwidth]
    {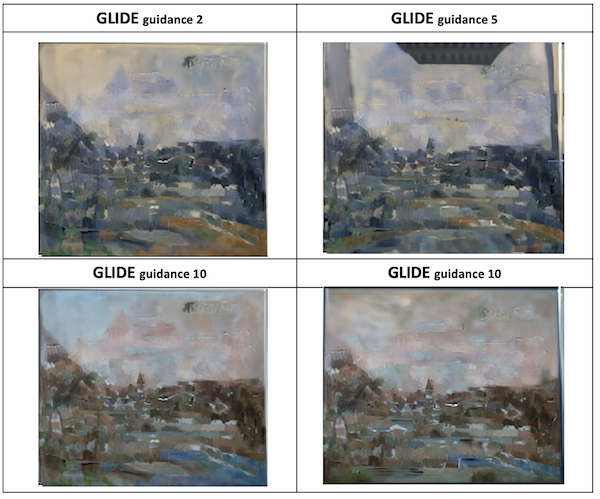}
  \caption{Comparison of inpainting models for Cezanne's unfinished "Turning Road" (1905). GLIDE's output is presented with several values for the Guidance Scale. Image from Wiki Commons.}
  \label{fig:Cezanne-comparison2}
\end{figure}

The panels on \ref{fig:Cezanne-comparison2} show Cezanne's  unfinished "Turning Road" (1905), which has whole sections of the canvas bare. The inpainting of this work is more open-ended given the style of the painter. For this reason, all models present equivalent performance, to the casual eye. To generate GLIDE's prompt we did a crowd-sourcing experiment and relied on the expertise of ten visual artist from EleutherAI who suggested by consensus  "the fog of the valley painting from last century". The two GLIDE images on the bottom were generated by using different seeds on the same prompt.

{\small
\bibliographystyle{ieee_fullname}
\bibliography{egbib}
}

\end{document}